\documentclass{sigchi-ext}
\usepackage[T1]{fontenc}
\usepackage{textcomp}
\usepackage[scaled=.92]{helvet} 
\usepackage{graphicx} 
\usepackage{balance}  
\usepackage{booktabs} 
\usepackage{ccicons}  
\usepackage{ragged2e} 
\usepackage{hyperref}
\usepackage{tikz}
\usetikzlibrary{shapes.geometric, arrows}

\tikzstyle{start} = [rectangle, rounded corners, minimum width=3cm, minimum height=1cm,text centered, draw=black, fill=red!30]
  \tikzstyle{process} = [rectangle, minimum width=3cm, minimum height=1cm, text centered, draw=black, text width=3cm, fill=orange!30]
\tikzstyle{decision} = [diamond, minimum width=2.5cm, minimum height=1.5cm, text badly centered, draw=black, fill=green!30, aspect=2]
\tikzstyle{stop} = [trapezium, trapezium left angle=70, trapezium right angle=110, trapezium stretches=false, text centered, draw=black, fill=blue!30, text width=2.5cm]

\tikzset{
    >=stealth',
    punkt/.style={
           rectangle,
           rounded corners,
           draw=black, very thick,
           text width=6.5em,
           minimum height=2em,
           text centered},
    pil/.style={
           ->,
           thick,
           shorten <=2pt,
           shorten >=2pt,}
}

\def\plainkeywords{Fairness and Bias in Artificial Intelligence; Ethics of AI}

\title{Getting Fairness Right: Towards a Toolbox for Practitioners}

\numberofauthors{3}
\author{%
  \alignauthor{%
    \textbf{Boris Ruf}\\
    \affaddr{Research Data Scientist} \\
    \affaddr{AXA, REV Research} \\
    \affaddr{Paris, France} \\
    \email{boris.ruf@axa.com} } \vfil \alignauthor{%
    \textbf{Chaouki Boutharouite}\\
    \affaddr{Research Manager} \\
    \affaddr{AXA, REV Research}\\
    \affaddr{Paris, France}\\
    \email{chaouki.boutharouite@axa.com} } \vfil \alignauthor{%
    \textbf{Marcin Detyniecki}\\
    \affaddr{Head of R\&D} \\
    \affaddr{AXA, REV Research}\\
    \affaddr{Paris, France}\\
    \email{marcin.detyniecki@axa.com} }}

\begin{document}

\CopyrightYear{2020}
\setcopyright{rightsretained}
\conferenceinfo{CHI'20,}{April  25--30, 2020, Honolulu, HI, USA}
\isbn{978-1-4503-6819-3/20/04}
\doi{https://doi.org/10.1145/3334480.XXXXXXX}
\copyrightinfo{\acmcopyright}

\maketitle

\RaggedRight{} 


\begin{abstract}
The potential risk of AI systems unintentionally embedding and reproducing bias has attracted the attention of machine learning practitioners and society at large.  As policy makers are willing to set the standards of algorithms and AI techniques, the issue on how to refine existing regulation, in order to enforce that decisions made by automated systems are fair and non-discriminatory, is again critical. Meanwhile, researchers have demonstrated that the various existing metrics for fairness are statistically mutually exclusive and the right choice mostly depends on the use case and the definition of fairness.

Recognizing that the solutions for implementing fair AI are not purely mathematical but require the commitments of the stakeholders to define the desired nature of fairness, this paper proposes to draft a toolbox which helps practitioners to ensure fair AI practices. Based on the nature of the application and the available training data, but also on legal requirements and ethical, philosophical and cultural dimensions, the toolbox aims to identify the most appropriate fairness objective. This approach attempts to structure the complex landscape of fairness metrics and, therefore, makes the different available options more accessible to non-technical people. In the proven absence of a silver bullet solution for fair AI, this toolbox intends to produce the fairest AI systems possible with respect to their local context.
\end{abstract}




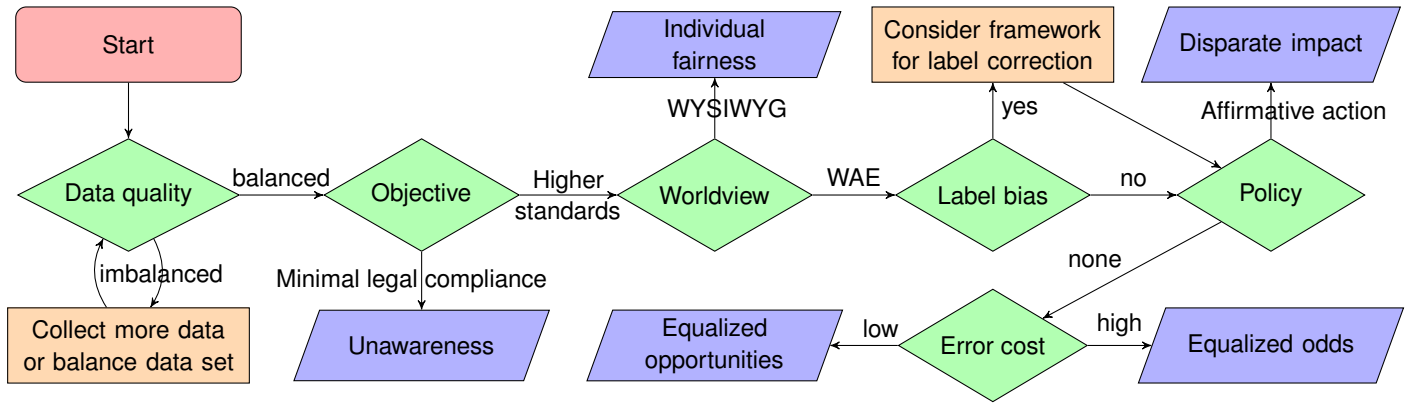
\begin{figure*}[hbt!]
  \centering
  \begin{tikzpicture}[node distance=2cm]
  
  \node (start) [start] {Start};
  \node (data) [decision, below of=start, align=center] {\hyperref[ssec:data_quality]{Data quality}};
  \node (collect) [process, below of=data] {Collect more data or balance data set};
  \node (objective) [decision, right of=data, xshift=1.9cm] {\hyperref[ssec:objective]{Objective}};
  \node (worldview) [decision, right of=objective, xshift=1.9cm] {\hyperref[ssec:worldview]{Worldview}};
  \node (label_bias) [decision, right of=worldview, xshift=1.7cm] {\hyperref[ssec:label_bias]{Label bias}};
  \node (correction) [process, above of=label_bias, xshift=0cm] {\hyperref[ssec:label_correction]{Consider framework\\ for label correction}};
  
  \node (unawareness) [stop, below of=objective, inner ysep=0.35cm] {Unawareness};

  \node (policy) [decision, right of=label_bias, xshift=1.7cm] {\hyperref[ssec:policy]{Policy}};
  \node (individual_fairness) [stop, above of=worldview, inner ysep=0.11cm] {Individual fairness};
  \node (cost) [decision, below of=label_bias] {\hyperref[ssec:error_cost]{Error cost}};
  \node (equalized_opportunities) [stop, below of=worldview, inner ysep=0.11cm] {Equalized\\opportunities};
  \node (equalized_odds) [stop, below of=policy,inner xsep=0.15cm, inner ysep=0.35cm] {Equalized odds};
  \node (disparate_impact) [stop, above of=policy, inner ysep=0.35cm] {Disparate impact};
  
  \path[->] (data) edge [bend left] node[right=-0.95cm] {imbalanced} (collect);
  \path[->] (collect) edge [bend left] node {} (data);
  \path[->] (worldview) edge [] node[right=-0.75cm] {WYSIWYG} (individual_fairness);
  \path[->] (worldview) edge [] node[above] {WAE} (label_bias);
  \path[->] (policy) edge [] node[right=-1.05cm] {Affirmative action} (disparate_impact);
  \path[->] (policy) edge [pos=0.4] node[right=-1.2cm] {none} (cost);
  \path[->] (cost) edge [pos=0.25] node[above] {low} (equalized_opportunities);
  \path[->] (cost) edge [] node[above] {high} (equalized_odds);
  \path[->] (objective) edge [] node[right=-2.05cm]{Minimal legal compliance} (unawareness);
  \path[->] (objective) edge [] node[align=center]{Higher\\standards} (worldview);
  \path[->] (start) edge [] node {} (data);
  \path[->] (data) edge [] node[above] {balanced} (objective);
  \path[->] (label_bias) edge [] node[above]{no} (policy);
  \path[->] (label_bias) edge [] node[right]{yes} (correction);
  \path[->] (correction) edge [] node[right]{} (policy);
 
  \end{tikzpicture}
  \caption{Proposal of a toolbox to structure the existing landscape of most commonly used fairness definitions. The diamonds mark decision points, rectangles symbolize actions, and trapezes represent the suggested fairness objective. }
  \label{fig:flow_chart}
\end{figure*}

\section{Introduction}
Whilst AI systems provide promising new opportunities for society and technology users, they may also introduce new and changed risks. Among these, AI systems can potentially reproduce and reinforce bias~\cite{DBLP:journals/corr/abs-1301-6822, DBLP:journals/corr/BolukbasiCZSK16a, DBLP:journals/corr/IslamBN16, 2017arXiv170309207B}. The research community has proposed different mathematical definitions of fairness and many, mostly technical mitigation strategies~\cite{mehrabi2019survey}. However, as there is no such thing as one uniform notion of fairness, obtaining a general kind of fairness by optimizing all metrics simultaneously was shown to be impossible~\cite{Friedler2016,Corbett-Davies2018}. Still, fixing biases in algorithms remains a technical problem which is far easier to solve than correcting cognitive bias.  Therefore, it is now about explaining the available fairness objectives to the broader audience, defining the appropriate objectives depending on a given context, and matching them with the findings of the fair machine learning research community.

\marginpar{%
  \vspace{-247pt} \fbox{%
    \begin{minipage}{0.925\marginparwidth}
      \textbf{Example of jurisdiction: Fairness in the EU} \\
      \vspace{1pc} When it comes to non-discrimination, the \emph{Convention for the Protection of Human Rights and Fundamental Freedoms} defines the "Prohibition of discrimination" in Article 14~\cite{Europe:1950uq}. This principle is further contained in the \emph{Charter of Fundamental Rights of the European Union} which states in Article 21 that "[a]ny discrimination based on any ground such as sex, race, colour, ethnic or social origin, genetic features, language, religion or belief, political or any other opinion, membership of a national minority, property, birth, disability, age or sexual orientation shall be prohibited."~\cite{EUCharter}
      
    \end{minipage}}\label{sec:sidebar} }

\section{Toolbox}
The flow chart in Figure~\ref{fig:flow_chart} shows the proposed toolbox to identify the most adapted fairness definition depending on the expected notion of fairness and the application scenario. The decision points require the stakeholder to assess the application and to commit to the desired fairness characteristics. In the following, each of such steps is explained in further detail.

\subsection{Data quality}\label{ssec:data_quality}
A major source for unwanted bias in machine learning algorithms is the training data. To the model, this data corresponds as the ground truth. If it does not confidently represent the real distribution, the model is likely to be exposed to selection bias. This will result in discriminating outcome, but also affect the accuracy of the model.

\subsection{Objective}\label{ssec:objective}
Most legal frameworks are providing clear guidelines on data privacy. In the EU, for example, the collection and use of sensitive personal data are strictly regulated by the General Data Protection Regulation (GDPR). More specifically, the use and processing of a list of sensitive attributes is forbidden. 

Depending on the applicable law, the principles of fairness and non-discrimination (see sidebar on the left) may already be met when the sensitive attributes are omitted. However, due to underlying correlations in big data and the nature of pattern identifying machine learning algorithms, this requirement may not be sufficient to achieve fair outputs. When the goal is to pursue higher standards based on ethics guidelines or to anticipate future changes in law, it will take further, more profound actions.

\subsection{Label bias}\label{ssec:label_bias}
Machine learning algorithms are trained by example. The assumption is that the labels of the training data are correct, they constitute the supposed ground truth. Depending on the data set, this can be guaranteed, for example when the labels result from objective measurements (e.g. by a thermometer) or describe indisputable facts (e.g. the borrower did or did not reimburse the loan). However, when labels represent historical human decisions, they may as well contain human bias. As the labels serve as reference to estimate the model's accuracy but also to satisfy a fairness metric when this one is based on classification rates, it is crucial to mitigate this potential source of bias, possibly using a label correction framework~\cite{NIPS2019_9082, Jiang2019}.

\subsection{Worldview}\label{ssec:worldview}
There is no such thing as one uniform notion of fairness, therefore \cite{Friedler2016} formalized two opposing worldviews. Settling for one view helps narrow down the number of appropriate fairness objectives. The worldview \emph{what you see is what you get (WYSIWYG)} assumes the absence of structural bias in the data. This view assumes that any statistical variation in different groups actually represents deviating base rates which should get explored. On the other hand, the worldview \emph{we're all equal (WAE)} presupposes equal base rates for all groups. Possible deviations are considered as unwanted structural bias that needs to get corrected.

\marginpar{%
  \vspace{-287pt} \fbox{%
    \begin{minipage}{0.925\marginparwidth}
      \textbf{Fairness objectives} \\
      \vspace{1pc} \textbf{Unawareness:} Remove sensitive attributes from the data set. Fairness through data sanitization~\cite{Pedreshi2008}.\\
      \vspace{1pc} \textbf{Individual fairness:} Similar individuals should be treated similarly based on an adequate distance metric~\cite{Dwork2012}.\\
      \vspace{1pc} \textbf{Disparate impact:} Minimize the absolute difference of outcome distributions of all groups~\cite{Feldman2014}.\\
      \vspace{1pc} \textbf{Equalized odds:} Optimize towards equal positive and negative classification rates across all groups~\cite{Hardt2016}.\\
      \vspace{1pc} \textbf{Equalized opportunities:} Optimize towards equal positive classification rates across all groups~\cite{Hardt2016}.\\
      
    \end{minipage}}\label{sec:sidebar} }
    
\subsection{Policy}\label{ssec:policy}
Fairness objectives can go beyond equal treatment of different groups or similar individuals. If the target is to bridge prevailing inequalities by boosting underprivileged groups, affirmative actions or quotas can be valid measures. Committing to a goal like this results in subordinating the algorithm's accuracy to such a policy's overarching goal.

\subsection{Error cost}\label{ssec:error_cost}
Depending on the use case, the consequences of misclassification can range from minor problems (e.g. inappropriate movie recommendation) to life-affecting ones (e.g. bail and parole decisions). For high-risk applications, the goal is to keep positive and negative classification rates equal for all groups. For low-risk applications the fairness objective could be weakened by accepting a manageable degree of extra risk in order to increase utility of the metric~\cite{Hardt2016}.

\section{Conclusion}\label{ssec:conclusion}
Hence, we propose a toolbox for choosing the right fairness objective for the given use case. To achieve this, we identified crucial decision points and actions which support practitioners to make an informed selection when building fair AI systems. With this contribution, we hope to bring forward the development of actionable AI guidelines, going into real world practice. 
\balance{} 

\bibliographystyle{SIGCHI-Reference-Format}
\bibliography{sample}


\begin{thebibliography}{00}


\ifx \showCODEN    \undefined \def \showCODEN     #1{\unskip}     \fi
\ifx \showDOI      \undefined \def \showDOI       #1{{\tt DOI:}\penalty0{#1}\ }
  \fi
\ifx \showISBNx    \undefined \def \showISBNx     #1{\unskip}     \fi
\ifx \showISBNxiii \undefined \def \showISBNxiii  #1{\unskip}     \fi
\ifx \showISSN     \undefined \def \showISSN      #1{\unskip}     \fi
\ifx \showLCCN     \undefined \def \showLCCN      #1{\unskip}     \fi
\ifx \shownote     \undefined \def \shownote      #1{#1}          \fi
\ifx \showarticletitle \undefined \def \showarticletitle #1{#1}   \fi
\ifx \showURL      \undefined \def \showURL       #1{#1}          \fi

\bibitem{2017arXiv170309207B}
{Richard Berk}, {Hoda Heidari}, {Shahin Jabbari}, {Michael Kearns}, {and}
  {Aaron Roth}. 2017.
\newblock \showarticletitle{Fairness in Criminal Justice Risk Assessments: The
  State of the Art}.
\newblock {\em Sociological Methods \& Research\/} (Mar 2017).
\newblock


\bibitem{DBLP:journals/corr/BolukbasiCZSK16a}
{Tolga Bolukbasi}, {Kai{-}Wei Chang}, {James~Y. Zou}, {Venkatesh Saligrama},
  {and} {Adam Kalai}. 2016.
\newblock \showarticletitle{Man is to Computer Programmer as Woman is to
  Homemaker? Debiasing Word Embeddings}.
\newblock {\em CoRR\/}  {abs/1607.06520} (2016).
\newblock


\bibitem{Corbett-Davies2018}
{Sam Corbett{-}Davies} {and} {Sharad Goel}. 2018.
\newblock \showarticletitle{The Measure and Mismeasure of Fairness: {A}
  Critical Review of Fair Machine Learning}.
\newblock {\em CoRR\/}  {abs/1808.00023} (2018).
\newblock


\bibitem{Dwork2012}
{Cynthia Dwork}, {Moritz Hardt}, {Toniann Pitassi}, {Omer Reingold}, {and}
  {Richard~S. Zemel}. 2011.
\newblock \showarticletitle{Fairness Through Awareness}.
\newblock {\em CoRR\/}  {abs/1104.3913} (2011).
\newblock


\bibitem{EUCharter}
{{European Union}}. 2010.
\newblock {\em Charter of Fundamental Rights of the European Union}. Vol.~53.
\newblock European Union, Brussels. 380 pages.
\newblock


\bibitem{Feldman2014}
{Michael Feldman}, {Sorelle Friedler}, {John Moeller}, {Carlos Scheidegger},
  {and} {Suresh Venkatasubramanian}. 2014.
\newblock \showarticletitle{{Certifying and removing disparate impact}}.
\newblock  (2014), 1--28.
\newblock
\showISBNx{9781450336642}
\showISSN{1083-4389}


\bibitem{Friedler2016}
{Sorelle~A. Friedler}, {Carlos Scheidegger}, {and} {Suresh Venkatasubramanian}.
  2016.
\newblock \showarticletitle{{On the (im)possibility of fairness}}.
\newblock  (Sep 2016).
\newblock


\bibitem{Hardt2016}
{Moritz Hardt}, {Eric Price}, {and} {Nathan Srebro}. 2016.
\newblock \showarticletitle{Equality of Opportunity in Supervised Learning}.
\newblock {\em CoRR\/}  {abs/1610.02413} (2016).
\newblock


\bibitem{DBLP:journals/corr/IslamBN16}
{Aylin~Caliskan Islam}, {Joanna~J. Bryson}, {and} {Arvind Narayanan}. 2016.
\newblock \showarticletitle{Semantics derived automatically from language
  corpora necessarily contain human biases}.
\newblock {\em CoRR\/}  {abs/1608.07187} (2016).
\newblock


\bibitem{Jiang2019}
{Heinrich Jiang} {and} {Ofir Nachum}. 2019.
\newblock \showarticletitle{Identifying and Correcting Label Bias in Machine
  Learning}.
\newblock {\em CoRR\/}  {abs/1901.04966} (Jan 2019).
\newblock


\bibitem{mehrabi2019survey}
{Ninareh Mehrabi}, {Fred Morstatter}, {Nripsuta Saxena}, {Kristina Lerman},
  {and} {Aram Galstyan}. 2019.
\newblock A Survey on Bias and Fairness in Machine Learning.
\newblock   (2019).
\newblock


\bibitem{Europe:1950uq}
{Council of Europe}. 1950.
\newblock \showarticletitle{Convention for the Protection of Human Rights and
  Fundamental Freedoms}, Vol. Rome, 4.XI.
\newblock


\bibitem{Pedreshi2008}
{Dino Pedreshi}, {Salvatore Ruggieri}, {and} {Franco Turini}. 2008.
\newblock \showarticletitle{{Discrimination-aware data mining}}.
\newblock {\em Proceeding of the 14th ACM SIGKDD international conference on
  Knowledge discovery and data mining - KDD 08\/} (2008), 560.
\newblock
\showISBNx{9781605581934}
\showISSN{0309-0167 (Print)}


\bibitem{DBLP:journals/corr/abs-1301-6822}
{Latanya Sweeney}. 2013.
\newblock \showarticletitle{Discrimination in Online Ad Delivery}.
\newblock {\em CoRR\/}  {abs/1301.6822} (2013).
\newblock


\bibitem{NIPS2019_9082}
{Michael Wick}, {swetasudha panda}, {and} {Jean-Baptiste Tristan}. 2019.
\newblock \showarticletitle{Unlocking Fairness: a Trade-off Revisited}.
\newblock In {\em Advances in Neural Information Processing Systems 32},
  {H.~Wallach}, {H.~Larochelle}, {A.~Beygelzimer}, {F.~d\textquotesingle
  Alch\'{e}-Buc}, {E.~Fox}, {and} {R.~Garnett} (Eds.). Curran Associates, Inc.,
  8780--8789.
\newblock


\end{thebibliography}

\end{document}